\documentclass[times]{TRR}

\usepackage{moreverb,url}
\usepackage{multirow}

\usepackage[colorlinks,bookmarksopen,bookmarksnumbered,citecolor=red,urlcolor=red]{hyperref}

\usepackage[switch,pagewise,displaymath, mathlines]{lineno}

\newcommand\BibTeX{{\rmfamily B\kern-.05em \textsc{i\kern-.025em b}\kern-.08em
T\kern-.1667em\lower.7ex\hbox{E}\kern-.125emX}}

\def\volumeyear{2020}

\begin{document}

\runninghead{Karim et al.}

\title{Towards explainable artificial intelligence (XAI) for early anticipation of traffic accidents}

\author{Muhammad Monjurul Karim\affilnum{1}, Yu Li\affilnum{1} and Ruwen Qin\affilnum{1}}

\affiliation{\affilnum{1}Department of Civil Engineering, Stony Brook University, Stony Brook, NY 11794, USA\\}

\corrauth{Ruwen Qin, ruwen.qin@stonybrook.edu}

\begin{abstract}
Traffic accident anticipation is a vital function of Automated Driving Systems (ADSs) for providing a safety-guaranteed driving experience. An accident anticipation model aims to predict accidents promptly and accurately before they occur. Existing Artificial Intelligence (AI) models of accident anticipation lack a human-interpretable explanation of their decision-making. Although these models perform well, they remain a black-box to the ADS users, thus difficult to get their trust. To this end, this paper presents a Gated Recurrent Unit (GRU) network that learns spatio-temporal relational features for the early anticipation of traffic accidents from dashcam video data. A post-hoc attention mechanism named Grad-CAM is integrated into the network to generate saliency maps as the visual explanation of the accident anticipation decision. An eye tracker captures human eye fixation points for generating human attention maps. The explainability of network-generated saliency maps is evaluated in comparison to human attention maps. Qualitative and quantitative results on a public crash dataset confirm that the proposed explainable network can anticipate an accident on average  4.57 seconds before it occurs, with 94.02\% average precision. In further, various post-hoc attention-based XAI methods are evaluated and compared. It confirms that the Grad-CAM chosen by this study can generate high-quality, human-interpretable saliency maps (with 1.23 Normalized Scanpath Saliency) for explaining the crash anticipation decision. Importantly, results confirm that the proposed AI model, with a human-inspired design, can outperform humans in the accident anticipation.
\end{abstract}

\maketitle

\section{Introduction}

Autonomous driving research is advancing rapidly. Deep learning and computer vision are contributing technologies in this exciting journey \cite{janai2020computer, eskandarian2019research, muhammad2020deep}. While autonomous vehicles are blooming, there are still cases where autonomous vehicles are involved in crashes \cite{waymo_2020, ca_dmv_2021}. Accident anticipation and avoidance are essential safety functions required for autonomous vehicles. From the dashboard camera (dashcam) video, an accident anticipation model aims to predict if an accident would occur shortly. Specifically, the model classifies the driving scene of the near future as one with or without accident risk. This accident anticipation model is a desired safety-enhancement capability for not only autonomous vehicles but also countless human-driving vehicles. Successful anticipation of accidents from the widely deployed dashcams even just a few seconds ahead would effectively increase the situational awareness of human drivers, Advanced Driver Assistance Systems (ADAS), and autonomous vehicles to trigger a higher level of preparedness for accidents prevention.


Multiple computer vision-based deep learning models for the early anticipation of traffic accidents were developed recently with outstanding performance  \cite{bao2020uncertainty, you2020traffic, li2021crash, karim2021dynamic}. Each of the models may have millions of abstract parameters to learn from big data. Those learnable parameters are not directly attached to the physical nature of the problem to be solved. Therefore, despite the outstanding performance, the models' complex, black-box nature discourages societal acceptance. This issue is more acute for accident prediction models than many other Artificial Intelligence (AI) models. Accident anticipation is a high-stake, safety-critical function directly related to human lives. Thus, making AI models' decisions explainable to users who put their lives in the hands of AI is indispensable.



Importantly, humans' trust in autonomous driving technologies is a prerequisite for the mass adoption of autonomous vehicles \cite{ayoub2021modeling,shariff2017psychological,olaverri2020promoting,raats2020trusting,choi2015investigating, shen2020explain}. Shariff et al. \cite{shariff2017psychological} found that 78\% of Americans reported fearing to ride in an autonomous vehicle, and only 19\% indicated that they would trust autonomous vehicles. If people cannot verify the safety and reliability of autonomous driving technologies, the perceived trustworthiness of the technologies is compromised \cite{olaverri2020promoting}. Shariff et al. also concluded that further research needs to identify the information required to form trustable mental models of autonomous vehicles. Ha et al. \cite{ha2020effects} experimentally confirmed that explanations increase humans' trust in autonomous driving. Researchers have reached a consensus that building humans' trust in AI is an explanation of the decisions necessity. European Union has already made a legal regulation that people can request an explanation of AI decisions if they are affected by the decisions significantly \cite{goodman2017european}. Besides, the ability to explain the decision of an accident anticipation model would support insurance, legal, and regulatory entities to assess the model for the liability purpose \cite{zablocki2021explainability}. Therefore, it is imperative to make it transparent to people why AI models can anticipate traffic accidents. Achieving this goal will help remove the obstacle of integrating AI-powered accident anticipation into humans' daily lives.

Explainable AI (XAI) is a rapidly growing research topic that aims to develop algorithms and tools to generate high-quality, interpretable, intuitive, and human-understandable explanations of AI models' decisions. Saliency maps are among the XAI tools that visually explain the decision made by a computer vision-based deep neural network. Salient regions in an image are those firmly relevant to the AI model's decision. Various methods exist for creating high-quality, easily interpretable saliency maps \cite{simonyan2014deep, mahendran2016salient, montavon2017explaining, montavon2019layer, zhou2016learning, selvaraju2017grad, chattopadhay2018grad, fu2020axiom, muhammad2020eigen}. While these models have built a methodological foundation for XAI, their assessment is challenging. Questions remain open regarding the trustworthiness of the explanation generated by XAI. One possible method to assess the quality of an XAI tool or algorithm is to let humans provide additional annotation and then evaluate the match level of human annotation with the explanation produced by saliency maps. However, human annotation can be expensive to acquire. XAI has been receiving growing attention in autonomous driving. Attempts are noticed to explain the functions of various AI models for autonomous driving \cite{bojarski2018visualbackprop, kim2018textual, cultrera2020explaining, sauer2018conditional, gallitz2019interpretable, liu2020interpretable}. Yet, XAI studies have not caught the accelerating pace of AI-powered accident anticipation research.

This paper aims to develop an explainable deep neural network for early accident anticipation. The network comprises two major components. The first component is a Gated Recurrent Unit (GRU) network that analyzes the video captured by the dashcam to determine if an accident may occur shortly. This network mimics the human visual perception of accident risks during driving. Like the eyes of a driver, the dashcam captures the driving scene that contains comprehensive information of the surrounding. Similar to the brain of the driver, the deep neural network processes complex visual information to perceive the accident risk. The second component is the Gradient-weighted Class Activation Map (Grad-CAM) method that generates saliency maps to explain decisions made by the accident anticipation network. The saliency maps highlight pixels in a video that causally affect decisions made by the AI model. The generated explanation will help lift the psychological barrier for humans to take benefits promised by the AI-powered accident anticipation network. This study compares where the AI focuses with where the drivers look in predicting a future accident. This comparison reveals the quality of the saliency maps as a visual explanation.


In summary, contributions of this paper are three folds:
\begin{itemize}
\item development of a deep neural network that learns the spatio-temporal relationship among visual features embedded in dashcam videos to predict if a traffic accident would occur shortly,
\item integration of the developed deep network with the Grad-CAM method and its variants to product high-qualify saliency maps for interpreting the network's prediction, and
\item collection of human gaze points to assess the quality of the XAI methods, which are affordable and efficient.
\end{itemize}

The remainder of this paper is organized as follows. The next section summarizes the related literature. Then, the proposed methodology to create the explainable accident anticipation network is delineated. After that, the implementation details and experimental evaluation of the proposed method are discussed, followed by qualitative and quantitative result analyses. Finally, the conclusion and future work are summarized at the end of this paper. 

\section{Literature Review} 
\subsection{Saliency Maps as XAI Tools}

Saliency maps can explain the decision of a neural network by highlighting regions of input images where the network responds the most regards its decision. Creating saliency maps for computer vision-based deep networks has become an XAI approach. Simonyan et al. \cite{simonyan2014deep} introduced a gradient-based method to generate saliency maps for Convolutional Neural Networks (CNNs). Their approach visualizes regions relevant to a CNN's decision after one forward and one backward propagation. The gradients from the end of the network are backpropagated and projected onto input images. Still, the vanilla gradient calculation produces noisy saliency maps. Therefore, subsequent methods, such as Guided Backpropagation \cite{mahendran2016salient}, Deep Taylor \cite{montavon2017explaining}, and Layerwise Relevance Propagation (LRP) \cite{montavon2019layer}, were developed to create better saliency maps by modifying the back-propagation algorithm. 


Another line of research developed a method to create Class Activation Maps (CAM) that smooths saliency maps by localizing the class-specific regions in images \cite{zhou2016learning}. CAM requires removing the fully connected layers from the top of a trained CNN to put a global average pooling (GAP) layer followed by a single fully connected layer. Then, the saliency map is computed by summing up the activations of the last convolutional layer for individual output classes. However, a modification of the original architecture requires retraining the network. To overcome this drawback, Gradient-weighted CAM (Grad-CAM) generates high-quality saliency maps on the original network architecture without any modification \cite{selvaraju2017grad}. Recent extensions of Grad-CAM such as Grad-CAM++ \cite{chattopadhay2018grad}, XGrad-CAM \cite{fu2020axiom}, and Eigen-CAM \cite{muhammad2020eigen} are powerful XAI tools for visualizing the class-specific decision that deep neural networks make. 

\subsection{XAI for Autonomous Driving}

XAI is receiving growing attention in autonomous driving. For example, Bojarski et al. \cite{bojarski2018visualbackprop} proposed an activation-based method to backpropagate activations for obtaining smooth heat maps. This method for explaining CNN works in a real-time manner to be integrated by autonomous driving. Kim et al. \cite{kim2018textual} adopted an attention-based method to filter out non-salient image regions and display only those causally affecting the steering control of a stand-alone vehicle. Similarly, Cultrera et al. \cite{cultrera2020explaining} also used an attention model to visualize the perception of deep networks for autonomous driving. Autonomous driving has employed saliency to explain AI models for navigation, lane change detection, and driving behavior reasoning (e.g., hazard stop, or red light stop) \cite{liu2020interpretable}. All these methods achieved an impressive performance in explaining decisions that AI models made for critical tasks of autonomous driving. However, the explanation of deep networks for early anticipation of traffic accidents is less common. One XAI study partially related to accident anticipation is noticed, which attempts to explain object-induced actions of vehicles  \cite{xu2020explainable}. This study focuses on scene understanding, highlighting salient objects in the input images which potentially lead to a hazard. The network architecture selects potential objects from the region proposals proposed by Faster R-CNN \cite{ren2015faster} without considering spatio-temporal relational information. Hence, it disregards motion information that is an essential determinant of accidents.

\section{Methodology}

\begin{figure*}[!ht]
  \centering
  \includegraphics[width=6.5in]{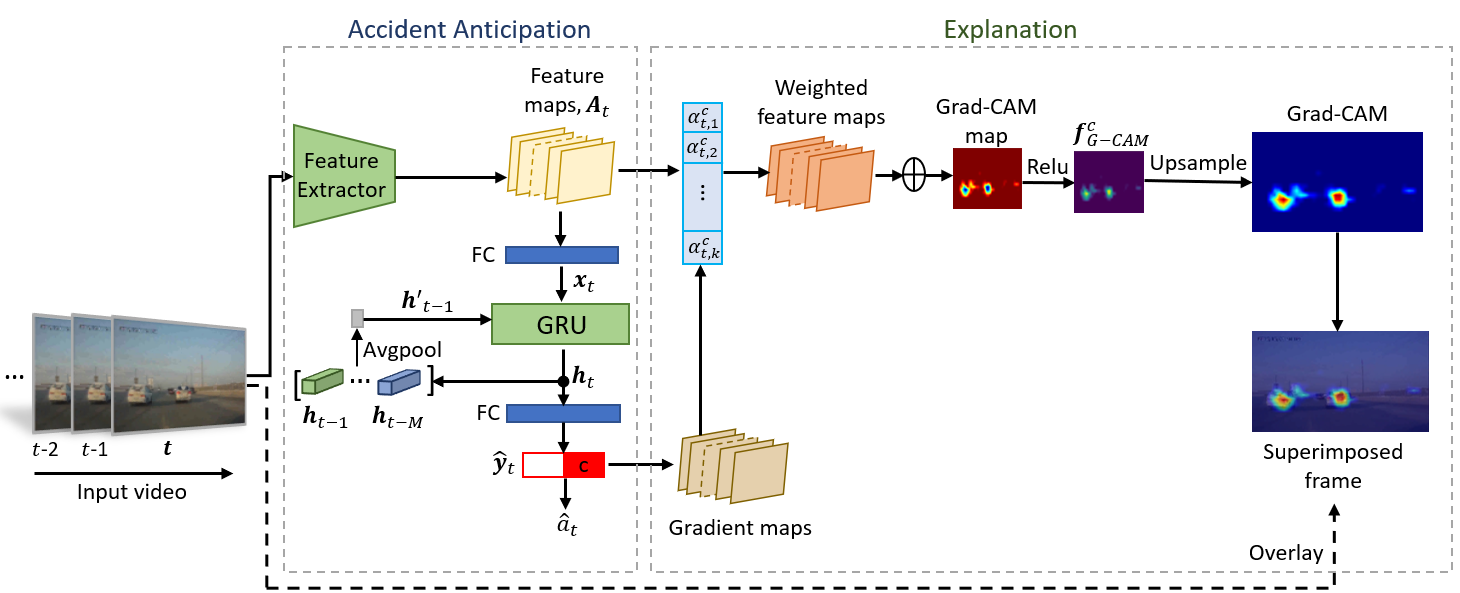}
  \caption{Overview of the proposed method to create the explainable accident anticipation network}
  \label{fig:overview}
\end{figure*}

Figure \ref{fig:overview} illustrates the proposed method to create the explainable accident anticipation network. The video captured by the dashcam, as a sequence of frames indexed by $t$, flows into a feature extractor. The feature extractor extracts a feature map $\pmb{A}_t$ from frame $t$. After passing a dense layer, the feature map becomes a feature vector, $\pmb{x}_t$, which goes to a Gated Recurrent Unit (GRU) to learn the hidden representation of the frame, $\pmb{h}_t$. Using the hidden representation, the network predicts the scores of the accident class $c$ and non-accident class $!c$, denoted by $\hat{\pmb{y}}_t=[\hat{y}_t^c,\hat{y}_t^{!c}]$. Gradient maps are calculated by backpropagating the prediction score with respect to the feature maps obtained from the feature extractor. Then, importance weights $\pmb{\alpha}^c$ are calculated from the gradients to aggregate all channels of the feature map into the Grad-CAM map that is further used to compute the final Grad-CAM saliency maps. Details of the method are discussed below.


\subsection{Feature Extraction} 
The base feature extractor that used in this study is the ResNet50 network initialized using parameters pre-trained on ImageNet \cite{krizhevsky2012imagenet}. The feature extractor extracts a feature map $\pmb{A}_t\in\mathbb{R}^{K\times U\times V}$ from frame $t$, for any $t$. That is, the feature map has $K$ channels, and the height and width of each channel are $U$ and $V$, respectively. Then, the flattened feature maps are passed through a dense layer to become a $d$ dimensional feature vector, $\pmb{x}_t\in\mathbb{R}^d$.



\subsection{Spatio-temporal Relational Learning with GRU}
Recurrent Neural Network (RNN) is a powerful tool for spatio-temporal sequential learning. Spatio-temporal relationships among image features contain important cues for accident anticipation. This study uses GRU, a particular type of RNN, to learn spatio-temporal relationships among the features by updating the hidden representation of each frame, $\pmb{h}_t$. GRU has two gates, a reset gate $\pmb{g}_t^{(r)}$ and an update gate $\pmb{g}_t^{(u)}$, which retain the most relevant information from the video sequence by filtering out irrelevant information. The data flowing through the GRU are expressed mathematically in equations (\ref{eq:resetgate}-\ref{eq:ht}):
\begin{equation}  
\pmb{g}_t^{(r)} = \sigma(\pmb{W}_g^{(r)}\pmb{x}_t + \pmb{B}_g^{(r)}\pmb{h}'_{t-1}),
\label{eq:resetgate}
\end{equation} 

\begin{equation} 
\pmb{r}_t =  \tanh (\pmb{W}_r \pmb{x}_t+\pmb{B}_r(\pmb{g}_t^{(r)} \circ \pmb{h}'_{t-1}) ), 
\end{equation}

\begin{equation} 
\pmb{g}_t^{(u)} = \sigma(\pmb{W}_g^{(u)}\pmb{x}_t + \pmb{B}_g^{(u)}\pmb{h}'_{t-1}), 
\end{equation} 

\begin{equation} 
    \pmb{h}_t = (1- \pmb{g}_t^{(u)}) \circ \pmb{r}_t  + \pmb{g}_t^{(u)} \circ \pmb{h}'_{t-1}, 
\label{eq:ht}
\end{equation} 
wherein $\sigma$ represents the sigmoid activation, $\circ$ is the element-wise product operator, $\pmb{W}$'s and $\pmb{B}$'s ($\in \mathbb{R}^{d\times d}$) are learnable parameters, and $\pmb{h}'_{t-1}$ is the average pooled hidden representations of the past $M$ frames:
\begin{equation} 
\pmb{h}^\prime_{t-1} =\mbox{avgpool}([\pmb{h}_{t-1}, \pmb{h}_{t-2}, \dots,  \pmb{h}_{t-M}]).
\end{equation} 
This study found that the average pooling of the most recent $M$ frames' hidden representations is better than the hidden representation of the last single frame in learning the contextual information.

Then, the hidden state $\pmb{h}_t$ is projected onto the prediction scores of the two classes - accident and no-accident - using a fully connected layer, $f_c$, with parameters $\pmb{W}_0$ and $\pmb{B}_0$ ($\in \mathbb{R}^{d\times d}$):
\begin{equation} 
\hat{\pmb{y}}_t =[\hat{y}_t^c,\hat{y}_t^{!c}]= f_c(\pmb{h}_t;\pmb{W}_0, \pmb{B}_0).
\label{eq:predictionscores}
\end{equation}

After that, the probability of seeing an accident shortly, predicted at time $t$, is obtained by the soft-max operation on $\hat{\pmb{y}}_t$:
\begin{equation}
    \hat{a}_t =\exp(\hat{y}^c)/[\exp(\hat{y}^c)+\exp{(\hat{y}^{!c}})]
\end{equation}

\subsection{Generating Saliency Maps with Grad-CAM}

This study uses Grad-CAM for generating saliency maps for the prediction of accident class $c$. Input video frames indexed by $t$ are fed to the network to calculate the Grad-CAM saliency maps for class $c$. Let $y^c_t$ be the score for class $c$ predicted at frame $t$, calculated in (\ref{eq:predictionscores}). First, the gradients of $y^c_t$ on the $k$th channel of the feature map $\pmb{A}_{t,k}$ are computed for all channels. Then, the gradients are averaged along the width and height of each channel to obtain the channel-wise importance weights $\alpha_{t,k}^c$:
\begin{equation}
    \alpha_{t,k}^c = \frac{1}{UV}\sum_{u=1}^{U}\sum_{v=1}^{V}\frac{\partial y^c_t}{\partial \pmb{A}_{t,k}(u,v)}.
\end{equation}
Afterwards, the importance weights are used to aggregate the $K$ channels of the feature map to get the Grad-CAM map. Finally, the activation function ReLU is applied to find the rectified Grad-CAM map for class $c$:
\begin{equation}
    \pmb{f}_{\text{G-CAM}}^c = ReLU\left(\sum_{k=1}^{K}\alpha_{t,k}^c \pmb{A}_{t,k} \right).
\end{equation}

The Grad-CAM map $\pmb{f}_{\text{G-CAM}}^c$ has the same spatial dimension as $\pmb{A}_{t,k}$. That is, ${\pmb{f}^c_{\text{G-CAM}}}\in\mathbb{R}^{U\times V}$. Its resolution is lower than that of the original video frame. Therefore, $\pmb{f}_{\text{G-CAM}}^c$ is up-sampled to have the same size of the input frame via the bilinear interpolation. The resized Grad-CAM map is superimposed with the input frame to explain the network's decision visually.

\section{Implementation and Experimental Evaluation}
\subsection{The Dataset}
A publicly available dataset called Car Crash Dataset (CCD) \cite{bao2020uncertainty} is used to train and evaluate the proposed accident anticipation network. CCD comprises videos that were captured by dashcams mounted on vehicles. The positive class of the dataset contains 1,500 videos that each contains an accident. The negative class has 3,000 videos without any accident. The dataset has diverse driving environment attributes. All the videos in CCD are trimmed to 5 seconds, and each video has 50 frames (i.e. 10 frames per second (fps)). The dataset is split into the training dataset with 3,600 videos and the testing dataset with 900 videos. 

\subsection{Model Training}
The training dataset, which comprises $N$ (=3,600) videos indexed by $n$, is used to train the network. The video-level label is $l_n=1$ if video $n$ is positive and $l_n=0$ if it is negative. Negative videos use the vanilla cross entropy loss function. Positive videos use an exponential cross entropy loss function to encourage early anticipation of accidents. That is, the loss for a positive video is near zero when time is far away from the accident, increases gradually as time is approaching the accident occurrence, and reaches the same level for negative videos at the accident occurrence and onward. Finally, the total loss on the training dataset is:
\begin{equation}
\begin{aligned}
    \mathcal{L} =&\sum_{n=1}^N\left[ -(1-l_n)\sum_{t=1}^T\log(1-\hat{a}_{t,n})\right. \\
    &\left.-l_n\sum_{t=1}^T\exp\left[-\max\left(\frac{\tau-t}{f},0\right)\right]\log(\hat{a}_{t,n})\right].
\end{aligned}
\label{eq:frame_loss}
\end{equation}
where $\hat{a}_{t,n}$ is the softmax probability that video $n$ belongs to the accident class, predicted at frame $t$.

The proposed network was trained and tested using an Nvidia V100 GPU with 32GB of memory. Input video frames were down sampled to $224 \times 224$ before feeding them to the network. The dimension of the feature map $\pmb{A}_{t}$ at the last convolution layer of the feature extractor is $14 (U) \times 14 (V)$ with 512 $(K)$ channels. The dimension of the feature vector before going to the GRU is 2,048 $(d)$. The dimension of hidden representations output from the GRU is 256. The number of hidden representations for average pooling is 3 $(M)$. The network was trained with the learning rate 0.0001, the batch size 10, and ReduceLROnPlateu as the learning rate scheduler. Adam optimizer was used to optimize the network for 30 epochs.

\subsection{Evaluation Metrics for the Accident Anticipation Model}

Assessment of the accident anticipation network was performed to determine how precisely and early the network can anticipate traffic accidents. Two metrics described below are used for the assessment.

\subsubsection{Average Precision}

On a testing video, if the softmax probability of accident, $\hat{a}_{t}$, exceeds a pre-specified threshold value $\bar{a}$ before an accident occurs, the video is predicted as positive. Otherwise, the prediction is negative. Accordingly, the recall (R) and precision (P) for the model were calculated based on the testing dataset:
\begin{equation}
    \mbox{R}=\frac{\# \mbox{true positive predictions}}{\# \mbox{positive videos}},
\end{equation}

\begin{equation}
    \mbox{P}=\frac{\# \mbox{true positive predictions}}{\# \mbox{positive predictions}}.
\end{equation}
Recall and precision values depend on the choice of classification threshold. A precision-recall curve can be created given various threshold values, and average precision (AP) is the area below this curve:
\begin{equation}
\mbox{AP}=\int \mbox{P}_{\mbox{\tiny{R}}}d\mbox{R}.
\end{equation}
where $\mbox{P}_{\mbox{\tiny{R}}}$ is the precision at a given recall value. AP measures how accurate the network is in anticipating accidents.

For a real-world implementation, the user may require a relatively high recall value. Hence, precision at the recall value 80\%, denoted by $P_{80\mbox{\tiny{R}}}$, was also calculated in this study.

\subsubsection{Time-to-Accident}
A positive video wherein the accident occurs at $\tau$ is predicted as positive when the probability $\hat{a}_{t}$ exceeds the classification threshold for the first time. The time-to-accident (TTA) is the from the time of prediction to the accident occurrence:
\begin{equation}
  \mbox{TTA} (\bar{a}) = \max\{\tau-t|\hat{a}_{t}>\bar{a}, 0\leq t\leq \tau\}.
\end{equation}
TTA measures how early the network can predict an accident, which is subject to the choice of the threshold value. The expected value of $\text{TTA}(\bar{a})$ is the mean TTA:
\begin{equation}
    \mbox{mTTA} = \mbox{E}_{\bar{a}}[\mbox{TTA}(\bar{a})].
\end{equation}
This study also calculates TTA$_{80\mbox{\tiny{R}}}$ in correspondence to P$_{80\mbox{\tiny{R}}}$.

\subsection{Creating Human Attention Maps as a Reference}

Drivers' visual attention is influenced by important visual cues in the traffic scene, such as a pedestrian, a cyclist, changes of traffic lights, or abnormal behavior of other vehicles. Therefore, drivers' gaze behavior can be used as the proxy of their attention. In this study, to evaluate the explainability of the developed network, human attention maps are computed and compared with the saliency maps generated by Grad CAM.

To obtain human attention maps, an eye tracking experiment was designed and performed using a Tobii Pro Fusion \cite{tobiiprofusion} eye tracker. Tobii Pro Fusion is a screen-based eye tracker. 12 volunteers participated in the experiment including 3 females and 9 males. The age of the participants ranges from 20 to 43. Their driving experience ranges from 3 months to 18 years. All the participants passed the eye tracker calibration test and thus are qualified for the eye tracking experiment. 

\begin{figure*}[!ht]
  \centering
  \includegraphics[width=\textwidth]{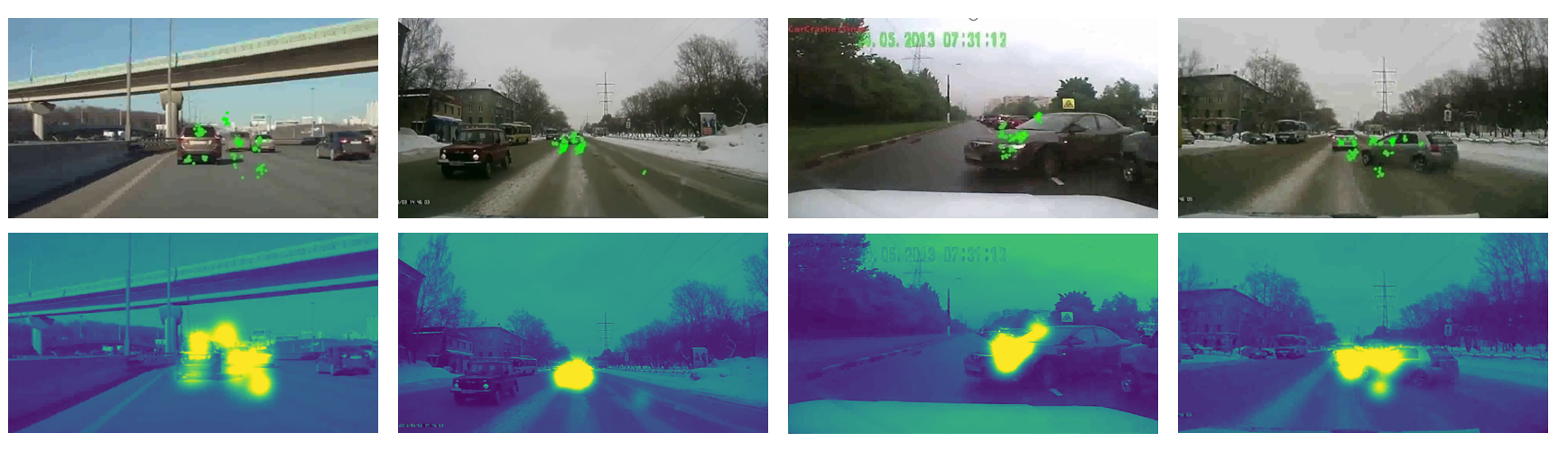}
  \caption{Human attention maps obtained from recorded gaze points} \label{fig:human_attention}
\end{figure*}

The experiment chose one hundred videos randomly from the test dataset of CCD to create an XAI test dataset. Fifty of the videos are positive videos, and the rest are negative videos. To assure the diversity of the scenes, the selected video clips contain a proportional number of environmental attributes. For example, 67\% of the video clips are normal weather and the remaining 33\% are snowy and rainy weather. These videos are in a random sequence so that participants do not know the class of the next video to watch. Furthermore, all the video clips are 5 seconds in length. For any positive video, the crash starting time is randomly placed in the last 2 seconds of the video clip. Due to the randomness, participants cannot predict the timing of crash occurrence from their experiences of watching other videos but the visual cues of the accident risk. Participants assume they are drivers when watching these videos. The eye tracker captures the participants' gaze data, including the timestamp and coordinates of each gaze point on the video frames. Each participant performed the experience once. In total, the experiment recorded about 720,000 gaze points, approximately 144 gaze points per frame. Gaze points could be classified into fixation, saccade, and unknown based on the angular speed of drivers' gaze movement.

A Gaussian filter of $30\times30$ pixels is used to convolute the count of gaze points on each frame into the human attention map. Figure \ref{fig:human_attention} shows a few samples of such human attention maps. The green dots in the first row of the figure are human gaze points. The second row shows the attention maps created after applying the Gaussian filter. 

While fixation points are usually used to create human attention maps, either licensed software is required, or a complex algorithm needs to be developed, to extract fixation points from gaze points. This study found that fixation points are the dominant class for drivers, counting 93\% of the collected gaze points. Therefore, it introduced a simplified method that uses all the gaze points to create human attention maps. While this approximation introduces a small amount of noise to the attention maps, objects that drivers attend to still stand out in the maps.

Obtained human attention maps are passed through a step filter to create fixation maps of binary pixel values. Pixels of value one in a fixation map are considered as fixated pixels.



\subsection{Evaluation Metrics for XAI}

This study evaluated the Grad-CAM method's ability to explain the prediction by the accident anticipation network by comparing the generated saliency maps to human fixation maps. The study considered both location-based and distribution-based metrics delineated below.


\subsubsection{Normalized Scanpath Saliency}

Normalized Scanpaht Saliency (NSS) measures the correspondence between saliency maps of the accident anticipation network and human fixation maps. Let $\hat{\pmb{S}}$ be the Grad-CAM saliency map of a frame and $\pmb{F}$ be the fixation map of the same frame. 
\begin{equation}
  \mbox{NSS}(\hat{\pmb{S}},\pmb{F}) = \frac{1}{\sum_iF_i} \sum_{i}^{}\frac{\hat{S}_i- \mu_{\hat{S}}}{\sigma_{\hat{S}}} F_i 
\end{equation}
where $i$ is the pixel index of $\hat{\pmb{S}}$ and $\pmb{F}$, $\hat{S}_i$ is the value of $\hat{\pmb{S}}$ at pixel $i$, and $F_i$ is the value of $\pmb{F}$ at pixel $i$. Mathematically, NSS is the average of normalized values at pixels of the saliency map where human fixations fall on. A positive NSS value means at a positive chance that the AI-rendered saliency map and the human fixation map have correspondence. The larger the NSS value, the stronger the correspondence.

\subsubsection{Area under ROC Curve}
The saliency map of a video frame can be seen as the prediction of the human fixation map of the frame. After applying a specific threshold value to the saliency map, the false positive rate and true positive rate of the binary classification result by the saliency map are obtained. The Receiver Operating Characteristic (ROC) curve can be created to measure the trade-off between true and false positives at various thresholds. Then, the Area Under the Curve (AUC) is computed to measure how well the model performs. Two different variants of AUC, AUC-J \cite{bylinskii2018different} and AUC-B \cite{bylinskii2018different}, are computed in this study.  The range of AUC is [0, 1]. The larger the AUC value, the higher chance that the saliency map and the human fixation map have correspondence. 

\subsubsection{Kullback-Leibler Divergence}
Kullback-Leibler (KL) divergence is generally used to estimate dissimilarity between two distributions. This study used KL to compare the network generated saliency maps with human fixation maps. Given a saliency map $\hat{\pmb{S}}$ and the corresponding human fixation map $\pmb{F}$, KL divergence can be approximated as:
\begin{equation}
    KL(\hat{\pmb{S}},\pmb{F})= \sum_{i}F_i \log\left(\epsilon + \frac{F_i}{\epsilon +\hat{S}_i}\right),
\end{equation}
where $\epsilon$ is a regularization constant. The range of KL divergence score is [0, $\infty$]. A low KL divergence score indicates a better approximation of the fixation map by the network generated saliency map.

\section{Results and Discussion}
\subsection{Performance of the Accident Anticipation Model}

The accident anticipation model was assessed on the test dataset of CCD. Two different experiments were conducted in this study by changing the number of hidden representations ($M$) fed to the GRU. During training, the network optimized the parameters by backpropagating the loss function. A set of trade-off solutions between AP and mTTA were obtained from different epochs of the training process. This paper only reports the results when the highest AP is achieved, shown in Table \ref{tab:accident_result}. 

\begin{figure*}[!ht]
  \centering
  \includegraphics[width=1\textwidth]{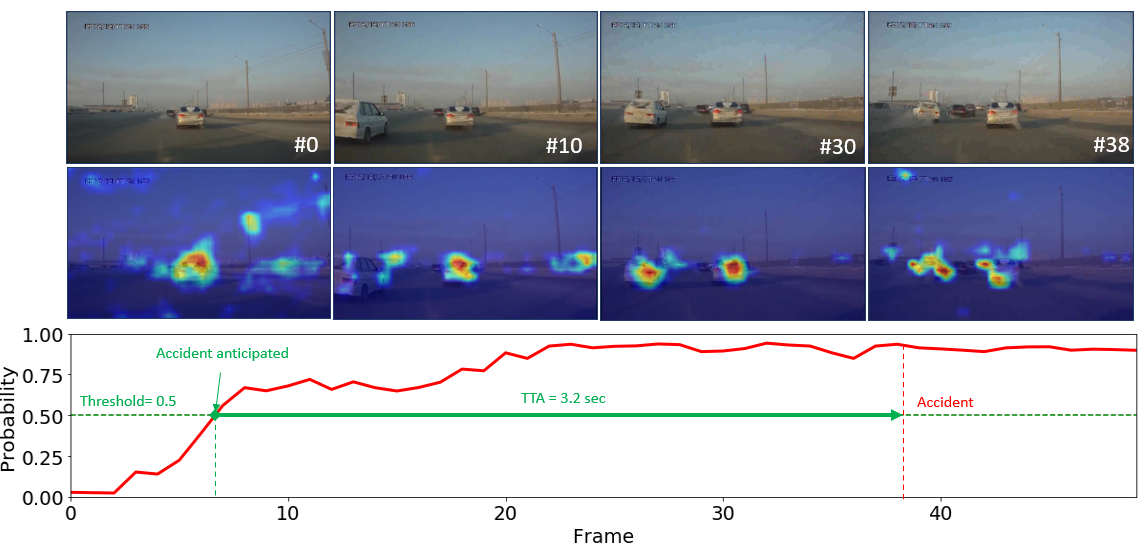}
  \caption{Visualizing the accident anticipation by the proposed network for explanation} \label{fig:accident_result}
\end{figure*}

\begin{table*}[!ht]
	\caption{Accident anticipation performance on CCD}\label{tab:dataset}
	\begin{center}
		\begin{tabular}{c| c| r |r| r| r}
			Experiment ID & $M$(\#) & AP(\%) & mTTA (s) & P$_{\tiny{\mbox{80R}}}$(\%) & TTA$_{\tiny{\mbox{80R}}}$ (s)\\\hline
			1   & 1 & 93.77 & 4.45 & 92.31 & 4.32\\
			2   & 3 & 94.02 & 4.57 & 93.02 & 4.50\\ \hline
		\end{tabular}
	\end{center}
	\label{tab:accident_result}
\end{table*}

In Table \ref{tab:accident_result}, the first experiment used the hidden representation of that last frame ($M=1$) to update the hidden representation of the current frame, whereas the second experiment mean-pooled the hidden representations of the last three frames ($M=3$). The result shows the integration of several recent frames’ hidden representations helps the network learn the temporal information better than using only a single frame. The achieved performance in the second experiment confirms that the proposed model can predict accidents very earlier (mTTA = 4.57 seconds and TTA$_{\mbox{\tiny{80R}}}$=4.50 seconds) with a very low false alarm rate (AP=94.02\% and P$_{\mbox{\tiny{80R}}}$=93.02\%).

Figure \ref{fig:accident_result} further illustrates an example of accident anticipation by the proposed network. In the figure, sample frames of a video clip are shown on the top, the corresponding saliency maps are in the middle, and the time series of the accident anticipation probability $\hat{a}_t$ denoted by the red colored curve is at the bottom. In the video, the accident starts at frame \#38. For the illustration purpose, a threshold value of 0.5 ($\bar{a}$) is set to trigger the prediction of an accident in this example. The probability of accident anticipation reached the threshold at frame \#7, which yields a TTA of 3.2 seconds. That is, the network predicted the accident 3.2 seconds before it happens. Hot spots in the saliency maps describe the network's concentration in the accident anticipation. Very early in the video (e.g., frame \#0), activations of the accident class were relatively sparse. Through learning the spatio-temporal relationships, the network concentrated more on objects that may involve in or be affected by an accident. In this example, two vehicles, A and B, in a relatively long distance from the ego-vehicle were first involved in a crash. Two additional vehicles, C and D, closer to the ego-vehicle were impacted by the crash. The one on the left adjacent lane (vehicle C) failed to avoid the crash, whereas the one in front (vehicle D) successfully avoided the accident. The saliency maps show that the network started attending to vehicles $C$ and $D$ very earlier (see frames \#10 and \#30). At frame \#38, the crash started with vehicles A and B that were involved first, and the saliency on vehicles C and D are split, and part of the saliency split apart goes towards vehicles A and B. 

\begin{figure*}[!ht]
  \centering
  \includegraphics[width=6.5in]{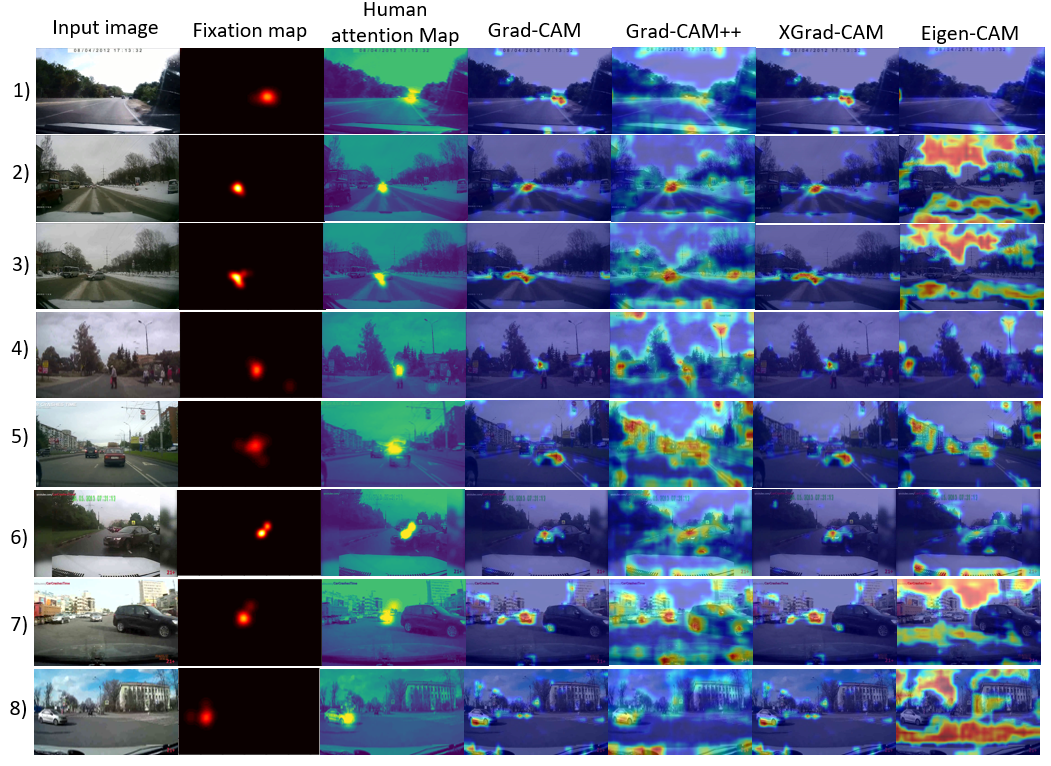}
  \caption{Visualizing the explanation generated by different XAI methods} \label{fig:xai_result}
\end{figure*}

\subsection{Performance of Explainable Artificial Intelligence (XAI) Methods}

\subsubsection{Qualitative Comparison}
Figure \ref{fig:xai_result} illustrates the qualitative results of several examples selected from the tested accident videos. The first column shows input video frames, the second column lists drivers' fixation maps. The third column is the human attention maps overlaid with the input image to show where drivers attend to. The last four columns, from the left to the right, are the saliency maps generated by Grad-CAM, Grad-CAM++, XGrad-CAM, and Eigen-CAM, respectively. The saliency maps are also overlaid with their respective input images to conveniently explain the accident anticipation decision made by the proposed accident anticipation network. All the selected sample frames are within 2 seconds before the accident occurs, where the accident involved or impacted objects are within the frames.

Humans have a natural tendency to focus on salient objects that have visual significance in understanding the situation. From the human attention maps (column 3 of Figure \ref{fig:xai_result}), it is evident that humans attend to vehicles or pedestrians that may be involved in or impacted by a traffic accident. The hottest spots in the saliency maps created by the Grad-CAM method (in column 4) closely overlap with the locations where humans fixate. This indicates the proposed accident anticipation network predicts high saliency values on the accident involved or impacted traffic agents and, accordingly, derives the prediction successfully.  Saliency maps created by XGrad-CAM (column 6) are very similar to those generated by Grad-CAM (column 4). Saliency maps obtained by Grad-CAM++ (Column 5) are sparser, wherein a lot of regions irrelevant to the accident are also considered as salient regions. Those are less capable of describing the network's decision. Similarly, Eigen-CAM creates very random saliency maps (column 7) that do not explain the network's decision well. 

The comparative analysis based on Figure \ref{fig:xai_result} reveals that the accident anticipation network developed in this study predicts a future accident by focusing on the most salient regions, just like a human does. The Grad-CAM method, and the XGrad-CAM method as well, can generate high quality saliency maps to explain how the proposed accident anticipation network makes a decision. Since the saliency maps generated by Grad-CAM and XGrad-CAM are interpretable to humans, they help users of the accident anticipation network establish their faith and trust in the underlying AI model.

\subsubsection{Quantitative Assessment}

The study further compared saliency maps generated by the XAI methods with human fixation maps using the evaluation metrics NSS, AUC-J, AUC-B, and KL divergence. For computing the NSS, a filter with a threshold value of 0.1 was applied to human attention maps to obtain binary fixation maps. Results for both the positive and the negative classes are summarized in Table \ref{tab:xai_result}. It should be noted that an XAI method is more favorable if it has larger values of NSS and AUC and a smaller value of KL divergence.
\begin{table*}[!ht]
	\caption{Performance comparison of XAI methods}\label{tab:dataset}
	\begin{center}
		\begin{tabular}{l| r| r |r| r|| r|r|r|r}
		  \hline
		  \multirow{2}{*}{{XAI methods}}& \multicolumn{4}{c||}{{Positive}} & \multicolumn{4}{c}{{Negative}} \\
		    \cline{2-9}
			 & {NSS} & {AUC-J} & {AUC-B} &  {KL} &  {NSS} &  {AUC-J} &  {AUC-B} &  {KL} \\
			 \hline
			 {Grad-CAM}  &  {1.23} &  {0.70} &  {0.65} &  {4.55} &  {0.50} &  {0.63} &  {0.60} &  {4.63} \\
			 {Grad-CAM++}  &  {1.13} &  {0.76} &  {0.75}  &  {12.75}  &  {0.86} &  {0.73} &  {0.72} &  {8.02}\\ 
			 {XGrad-CAM} &  {1.23} &  {0.70} &  {0.65} &  {4.55} &  {0.50} &  {0.62} &  {0.60} &  {4.63} \\
			 {Eigen-CAM} &  {0.10} &  {0.50} &  {0.49} &  {13.13} &  {0.04} &  {0.50}&  {0.48} &  {13.24}\\ \hline
		\end{tabular}
	\end{center}
	\label{tab:xai_result}
\end{table*}

It can be observed from Table  \ref{tab:xai_result} that Grad-CAM and XGrad-CAM achieve a similar performance in that they have near-identical values on all the metrics and on both classes. On the positive class, both of these two XAI methods receive reasonably good scores on all metrics. It should be mentioned that human fixation maps do not set an ideal performance benchmark for the XAI methods, but a reasonable reference. This is because humans attend to one spot at one time, and they may miss some critical salient regions in fast-changing driving scenes. Whereas the proposed AI network can attend to many salient regions or traffic agents in a video frame. Since human fixation maps, which are used as the ground-truth for assessing the XAI method, are not perfect saliency maps of accident risks, values of the metrics on the positive class were underestimated. The Grad-CAM++ method receives a lower NSS value ( {1.13}) but higher AUC-J value ( {0.76}) and AUC-B value ( {0.75}) on the positive class than both Grad-CAM and XGrad-CAM. The higher AUC values that the Grad-CAM++ method achieved on the positive class is because the AUC metrics do not penalize false positives whereas NSS does. Therefore, Grad-CAM++ achieves better AUC values although it generated sparse saliency maps that are visually different than human attention maps on the positive classes. Eigen-CAM performs poorly on all the four metrics on the positive class. The quantitative results suggest that the Grad-CAM method and the XGrad-CAM method can better explain the decision of the accident anticipation network by rendering high quality saliency maps on the positive class.

Performance of the XAI methods on the negative class was also evaluated in this study, with results presented on the right portion of Table \ref{tab:xai_result}. Values of the metrics on the negative class differ than the values on the positive class. Using the metric NSS as an example, the NSS value of any XAI method on the negative class is smaller than that on the positive class. This is because, when a driving scene is normal and low in the accident risk, the saliency maps that XAI methods generate for frames of the negative class are sparsely dispersed among different traffic agents or regions in the frames. However, humans always have a certain degree of fixation, regardless of the class of the driving scene. When the driving scene is normal, drivers fixate on regions and traffic agents for the purposes of navigating and conforming with driving rules. When the driving scene is risky, their attention is partially attracted by accident involved or impacted traffic agents. Therefore, on videos of the negative class, regions with human eye fixations do not have much saliency value for the accident risk. This explains the reason for obtaining low NSS values on the negative class. 

The NSS value of any XAI method on the negative class is below 1 and lower than the value on the positive class. Particularly, the between-class differences in the NSS values that Grad-CAM and XGrad-CAM obtain ( {1.23} vs.  {0.50}) is the largest among the four methods. The AUC values of Grad-CAM, Grad-CAM++, and XGrad-CAM on the negative classes are also lower than the values on the positive classes. Again, the between-class differences in AUC values that Grad-CAM and XGrad-CAM achieve are greater than Grad-CAM++. Compared to the KL divergence values on the positive class, the KL values of Grad-CAM, XGrad-CAM, and Eigen-CAM on the negative classes increase a little, whereas the value of Grad-CAM++ decreases. The comparison shows that Grad-CAM and XGrad-CAM are better than Grad-CAM++ and Eigen-CAM in differentiating risky driving scenes from normal scenes. The effectiveness of the metrics in a descending order is NSS, AUC, and KL.

\subsection{Inference Speed}
The inference speed of the proposed accident anticipation model is critical because accident anticipation needs to be real-time. This study thus evaluated the efficiency of the proposed method in terms of inference speed. By testing on the test dataset, it is found that the proposed method can process a video at a speed of 8.5 frames per second from loading a video frame to generating the prediction score. This study further evaluated the inference speed for the saliency map generation process. On average, Grad-CAM, Grad-CAM++, and XGrad-CAM take 11 ms, 13ms, and 12ms, respectively, for producing the saliency map for each video frame. However, Eigen-CAM takes a longer time, about 1.3 seconds per frame. 

\subsection{Challenges and Opportunities}

 {The output of the accident anticipation network developed in this paper is binary because the network classifies the driving scene of the near future as one with or without the accident risk. The current system can anticipate traffic accidents irrespective of the accident type. After sufficient training data become available, the current network can be extended to become a multiclass network to differentiate the accident type if an accident is anticipated to occur shortly. According to the Traffic Safety Facts Annual Report provided by National Highway Traffic Safety Administration (NHTSA), accidents resulting in fatalities, injuries, or property damages are associated with various types of first harmful events: 1) no collision with a motor vehicle in transport, 2) rear-end, 3) head-on, 4) angle, 5) sideswipe, and 6) unknown. During 2015~2019, 163,350 (96.65\%) fatal crashes in the United States are of the first four types \cite{TSFAR}. }


 {
Indeed, the forward-facing dashcam will not capture all types of accidents. For example, in a potential sideswipe accident, the vehicle to be collided on cannot capture the risk from its dashcam. However, dashcams are a low-cost sensor type widely deployed in many vehicles, not just one. The field of view limitation can be addressed well if most vehicles are equipped with dashcams and accessible to an accident anticipation system. For example, the vehicle likely to cause a sideswipe accident may see the risk from its dashcam. Other surrounding cars not to be involved in the accident may also perceive the sideswipe risk from their cameras. Likewise, in a potential rear-end accident, the car from the back can perceive the accident risk from its dashcam and thus try to avoid it, although the vehicle in front does not. Not to mention that adding additional camera sensors can increase the field of view. As we are moving towards the era of Vehicle-to-Vehicle (V2V) communication, vehicles that have anticipated an accident can share the information with others to alert them about possible accidents. The V2V communication will further broaden the impact of dashcam-based accident anticipation.}


\section{Conclusions}

This paper presented an explainable deep neural network for early anticipation of traffic accidents from dashcam videos. The proposed network is a Gated Recurrent Unit (GRU) that learns the spatio-temporal relationship between visual features of accidents by updating its hidden representations. Aggregating the hidden representations of multiple recent frames, the GRU learns the temporal-contextual information better, thus improving the network's performance. The experimental evaluation on the CCD dataset confirms that the proposed network can anticipate accidents very early (with 4.57-sec mTTA) and accurately (with 94.02\% AP). Additionally, the Grad-CAM is integrated into the proposed accident anticipation network to produce saliency maps that explain the network's decision visually. Human gaze data on a dashcam video dataset were collected to compare with the saliency maps generated by the explainable network. Four variants of XAI methods were further evaluated in this study. Out of these four methods, Grad-CAM and XGrad-CAM methods are most suitable because they generate high-quality visual explanation for the accident anticipation decision made by the network. Qualitative and quantitative evaluations both confirm that the proposed accident anticipation network has reliable and visually interpretable performance.  {Therefore, the proposed explainable network can not only build drivers' confidence in reliable AI models but address some drivers' blind trust in unreliable AI models.}

This paper has identified room for improvement. For example, the human subject experiment was performed in a laboratory setting. However, field testing scenarios can be more complex than in the controlled laboratory experiment, thus affecting human attention. Capturing human gaze data from actual driving conditions will help calibrate the results of evaluating the XAI methods. This study also found that the proposed AI method can surpass humans in finding salient regions to support the accident anticipation. Thus, to further strengthen humans’ trust in AI, an evaluation method will be developed to measure the strengths of AI over humans or vice versa. The accident anticipation network can enhance its performance further using human attention maps as a separate input channel.

 {Camera sensors have certain limitations, just like every other sensor type has its merits and limitations. For example, cameras may be blocked by dirt or mud while driving. Limitations of cameras do not restrict them from making positive contributions to transportation safety and autonomous driving. Moreover, it is unlikely that autonomous vehicles can rely on a single system to sense the environment and navigate themselves. This study envisions sensor fusion as a solution to address the limitations of individual sensor types.  Individual sensor technologies should be developed to their best to maximize the effectiveness of sensor fusion.  The advancement and synergy of these sensor technologies will positively contribute to the full driving automation.}

\begin{ac}
The authors confirm contribution to the paper as follows: study conception and design: M. M. Karim, R. Qin; data collection: Y. Li; analysis and interpretation of results: M. M. Karim, R. Qin; manuscript preparation: M. M. Karim, R. Qin. All authors reviewed the results and approved the final version of the manuscript.  
\end{ac}

\begin{dci}
The authors declared no potential conflicts of interest with respect to the research, authorship, and/or publication of this article. 
\end{dci}

\begin{funding}
All the authors received financial support from National Science Foundation through the award ECCS-2026357. Any opinions, findings, and conclusions or recommendations expressed in this material are those of the authors and do not necessarily reflect the views of the National Science Foundation.
\end{funding}

\begin{das}
Data, models, and code that support the findings of this study are available at \url{https://github.com/monjurulkarim/xai-accident}.
\end{das}

\bibliographystyle{TRR}
\bibliography{trb_template}





\end{document}